%% file: OpenInst.tex
\documentclass[preprint]{elsarticle}
\usepackage{lineno}
\modulolinenumbers[5]
\makeatletter
\def\ps@pprintTitle{%
  \let\@oddhead\@empty
  \let\@evenhead\@empty
  \let\@oddfoot\@empty
  \let\@evenfoot\@oddfoot
}
\makeatother

\usepackage[hidelinks,colorlinks,linkcolor=red]{hyperref}
\usepackage[utf8]{inputenc}
\usepackage[small]{caption}

\usepackage{times}
\usepackage{soul}
\usepackage{amsmath,amsfonts}
\usepackage{algorithm}
\usepackage{array}
\usepackage{textcomp}
\usepackage{stfloats}
\usepackage{url}
\usepackage{verbatim}
\usepackage{graphicx}
\usepackage{cuted}
\usepackage{natbib}

\usepackage{doi}
\usepackage{amssymb}
\usepackage{booktabs}
\usepackage{multirow}

\hypersetup{pdfauthor=author}

\begin{document}
\begin{sloppypar}

\begin{frontmatter}
\title{OpenInst: A Simple Query-Based Method for Open-World Instance Segmentation}



\author[hustaia,hustei]{Cheng Wang\fnref{1}}
\author[horizon]{Guoli Wang}
\author[horizon]{Qian Zhang}
\author[hustei]{Peng Guo\corref{corr}}\ead{guopeng@hust.edu.cn}
\author[hustei]{Wenyu Liu}
\author[hustei]{Xinggang Wang} 

\fntext[1]{This work was done when Cheng Wang was an intern at Horizon Robotics.}
\cortext[corr]{Corresponding author}
\address[hustaia]{Institute of Artificial Intelligence, Huazhong University of Science and Technology, Wuhan 430074, China}
\address[hustei]{School of Electronic Information and Communications, Huazhong University of Science and Technology, Wuhan 430074, China}
\address[horizon]{Horizon Robotics, Beijing 100086, China}

\begin{abstract}
\input{0-abs}
\end{abstract}

\end{frontmatter}

\section{Introduction}\label{sec:intro}
\input{1-intro}

\section{Related Work}\label{sec:related}
\input{2-related}

\section{Method}\label{sec:method}
\input{3-methods}

\section{Experiment}\label{sec:exp}
\input{4-exp}

\section{Conclusion}\label{sec:conclu}
\input{5-conclusion}

\section*{Acknowledgements}
This work was supported by the National Natural Science Foundation of China (NSFC) under grant No. 62276108.

\small
\bibliographystyle{plainnat}
\bibliography{6-sample-base}

\end{sloppypar}
\end{document}

%% file: 0-abs.tex


Open-world instance segmentation has recently gained significant popularitydue to its importance in many real-world applications, such as autonomous driving, robot perception, and remote sensing. However, previous methods have either produced unsatisfactory results or relied on complex systems and paradigms. We wonder if there is a simple way to obtain state-of-the-art results. Fortunately, we have identified two observations that help us achieve the best of both worlds: 1) query-based methods demonstrate superiority over dense proposal-based methods in open-world instance segmentation, and 2) learning localization cues is sufficient for open world instance segmentation. Based on these observations, we propose a simple query-based method named OpenInst for open world instance segmentation. OpenInst leverages advanced query-based methods like QueryInst and focuses on learning localization cues. Notably, OpenInst is an extremely simple and straightforward framework without any auxiliary modules or post-processing, yet achieves state-of-the-art results on multiple benchmarks. Specifically, in the COCO$\to$UVO scenario, OpenInst achieves a mask AR of 53.3, outperforming the previous best methods by 2.0 AR with a simpler structure. We hope that OpenInst can serve as a solid baselines for future research in this area.

%% file: 1-intro.tex
Object detection, which entails localizing and identifying objects within an image or a video sequence, is a crucial task in computer vision. Extensive researches have been conducted in this field, leading to the development of numerous works~\cite{ss,edgebox,rcnn,fast-rcnn,detr,owdetr,towod}. However, the majority of these works operate under the closed-world assumption, limiting their applicability to a pre-defined set of categories. Consequently, these models are inadequate when it comes to detecting novel objects that do not belong to the pre-defined categories, and their performance in recognizing such objects is limited. Direct deployment of these closed-world models in real-world scenarios can lead to serious consequences, such as false negatives that may cause accidents when applied to autonomous driving systems.


Open-world instance segmentation~\cite{oln,ggn,ldet,sois} has recently gained increasing attention for improving real-world applications. This task involves localizing and segmenting both seen and novel objects in an image, without necessarily recognizing them. This concept is in line with the occupancy network~\cite{occupancy} implemented in Tesla AI, which only considers the presence of objects and not their categories. It can be applied to downstream task open-vocabulary object detection and instance segmentation~\cite{ovvild,ovcm}, which aims to localize and segment both seen and unseen objects with recognizing their categories. The proposal process such as localize and segment can be realized by open-world instance segmentation methods. The recognition process is usually realized by vision-language models (VLMs) such as CLIP~\cite{clip} and ALIGN~\cite{align}.


Despite the recent advancements in open-world instance segmentation, it remains a challenging task, mainly due to the unlabeled objects in the training dataset. Traditional closed-world methods treat unlabeled objects as background, leading to overfitting of the labeled categories and poor generalization. LDET~\cite{ldet} proposes a simple-yet-effective data augmentation method to address the issue. It simply copies and pastes labeled objects onto background images. OLN~\cite{oln} takes a different approach by replacing the classification branch with a localization cues branch, which is not trained on negative (background) samples, thus improving the model's ability to generalize. Whereas, the performance of these methods is limited. Other methods, such as GGN~\cite{ggn}, GOOD~\cite{good}, and SOIS~\cite{sois}, have achieved better results but rely on complex systems and paradigms, making it difficult to compare their performance fairly. For example, GGN uses an auxiliary model to pseudo-annotate potential novel objects in the training images based on pairwise affinity. GOOD utilizes off-the-shelf models to produce depth and normal images of the original training images, which are then used to generate pseudo annotations. SOIS introduces an auxiliary branch to predict the foreground of an image and minimizes the difference between the outputs of the foreground branch and mask branch as a consistent loss during training.

The previous methods have failed to strike a balance between simplicity and high-performance. We seek an alternative approach that would allow us to achieve excellent results while maintaining a straightforward methodology. Recently, query-based detectors, such as DETR~\cite{detr}, have emerged as a promising option. Unlike dense proposal-based methods, query-based detectors employ N (e.g. 100) learnable queries to replace hundreds of thousands of pre-defined proposals, eliminating the need for many-to-one matching and post-processing.
We are drawn to query-based approach for two reasons. First, it offers a more concise structure than dense proposal-based methods. Second, it delivers better open-world instance segmentation performance than dense propose-based methods. In particular, we conducted a fair comparison of the class-agnostic QueryInst~\cite{queryinst} and Mask R-CNN~\cite{mask-rcnn} models in the COCO$\to$UVO scenarios and found that QueryInst outperformed Mask R-CNN by 4.9 mask AR.
Based on these findings, we develop our model, which we name OpenInst, using a query-based approach. We selected the advanced QueryInst~\cite{queryinst} as our baseline model for open-world instance segmentation, allowing OpenInst to maintain the simplicity of the base model while avoiding the complicated processing associated with dense proposals.


IoU-Net~\cite{iounet}, Mask Scoring R-CNN~\cite{msrcnn}, and FCOS~\cite{fcos} have demonstrated the effectiveness of geometric cues such as centerness, box IoU, mask IoU) in the closed-world tasks. OLN~\cite{oln} has further extended geometric cues to open-world proposal in dense proposal-based model. Those observations make it promising to extend geometric cues to query-based methods in terms of simplicity and performance. Therefore, OpenInst focuses solely on geometric cues in training. Going one step further, the geometric cues are derived from ground-truth boxes and masks without producing any extra information. Is it possible for the model to only learn to predict boxes and masks? Our study offers an affirmative answer.

Following prior works~\cite{oln,ggn,ldet}, we evaluate the generalizability of OpenInst through experiments in two major settings: cross-category and cross-dataset. 
The cross-category setting involves training the model on a pre-defined set of classes from a dataset and testing it on the remaining classes of the same dataset. The cross-dataset setting involves training the model on one dataset and testing it on a different dataset.

For cross-category generalizability evaluation, we use the COCO~\cite{coco} dataset. In this experiment, we train the model on the VOC~\cite{pascalvoc} classes of the training split and evaluate it on the remaining classes of the validation split. OpenInst achieves a mask AR of 28.2 in the VOC$\to$Non-VOC scenario, outperforming OLN by 1.3 AR. Additionally, OpenInst is compatible with the GGN method. And when powered by pseudo labels produced by GGN~\cite{ggn}, it achieves state-of-the-art results of 33.0 box AR and 30.1 mask AR in the VOC$\to$Non-VOC generalizability evaluation.
As for the cross-dataset evaluation, we conduct four experiments. We use COCO as the training set and test OpenInst on UVO~\cite{uvo}, Objects365~\cite{objects365}, and LVIS~\cite{lvis} datasets respectively. We have also trained OpenInst on Cityscapes~\cite{cityscapes} and evaluated it on Mapillary Vistas~\cite{mapillary}. OpenInst achieves state-of-the-art results in mask AR in all four scenarios.
In the COCO$\to$UVO scenario, In the COCO$\to$UVO scenario, OpenInst outperforms all dense proposal-based methods by a significant margin, achieving a mask AR of 53.3. When OpenInst is trained for 36 epochs, the mask AR of OpenInst can be further boosted to 53.3 AR, which outperforms SOIS~\cite{sois} by 2.0 AR. 
In the COCO$\to$Objects365, COCO$\to$LVIS, and Cityscapes$\to$Mapillary Vistas scenarios, OpenInst outperforms all other methods by a remarkable margin.
When trained without using geometric cues as the learning objective, OpenInst still performs well with only a 0.2 AR decrease from the state-of-the-art result in COCO$\to$UVO scenario. As the old saying goes, "Great Truths Are Always Simple". It also holds true here.


To summarize, our contributions can be concluded as follows:
\begin{itemize}
    \item OpenInst achieves state-of-the-art results on multiple datasets and significantly outperforms previous methods. Our study confirms that incorporating box IoU into query-based detectors improves their generalizability.
    \item To the best of our knowledge, OpenInst is the first work that achieves both simplicity and high performance, serving as a solid baseline for future research in the open-world community.
    \item We conduct extensive ablation studies to analyze each component used in our model. We hope those observations benefit future research in this area.
\end{itemize}

%% file: 2-related.tex
\subsection{Geometry Cues as Objectness}
\textbf{Learning-based geometry cues.}
Learning-based geometry objectness cues are utilized to evaluate the quality of the proposals and segments generated by a detector. These cues come in three popular forms: box intersection over union (IoU), mask intersection over union, and centerness. They have been integrated into dense proposal-based detectors serving as a learning objective as well as a ranking and filtering indicator.
IoU-Net~\cite{iounet} recognizes that classification scores can be unreliable in determining the quality of proposals. For example, proposals that only cover a partial but discriminative part of an object can still receive high classification scores. To address this issue, IoU-Net implements a box-IoU branch to estimate the IoU score of proposals and their corresponding ground-truth boxes. Similarly, Mask Scoring R-CNN~\cite{msrcnn} implements a mask-IoU branch to estimate the IoU score of predicted masks and their corresponding ground-truth masks. FCOS~\cite{fcos}, on the other hand, implements a centerness branch to estimate the centerness of the predicted boxes. Boxes with low centerness scores tend to be of low quality, and vice versa.
OLN~\cite{oln} takes this concept one step further by using geometry cues, such as IoU and centerness, solely as objectness supervision and proposals ranking indicators, thus improving the performance of the vanilla Faster R-CNN for open-world proposals. Similarly, GGN~\cite{ggn} leverages geometry objectness cues for learning-based ranking and pseudo-label generation. The use of pseudo labels with these cues has proven to be effective in improving performance in GGN.
OpenInst, on the other hand, uses geometry objectness cues for a different purpose - to solely serve as a learning objective and aid the box regression branch in localizing general objects, achieving both simplicity and high-performance.

\textbf{Learning-free geometry cues.}
Learning-free and heuristic geometry objectness cues have been the primary approaches used to determine the objectness of a proposal box. \citet{measuringobj} have used a combination of low-level geometry cues, such as salience, color contrast, and edge density, to estimate the objectness of a proposal. Selective Search (SS)~\cite{ss} has utilized up to 80 hierarchy grouping strategies to generate region candidates, with the objectness of each region being determined by the accumulation of positions within each hierarchy grouping strategy. EdgeBox~\cite{edgebox} has also utilized a simple approach, observing that the number of closed contours within a proposal box could represent its objectness.

\subsection{Query-based Detectors}
Dense proposal-based detectors~\cite{rcnn,fast-rcnn,faster-rcnn,fcos} often face difficulties in handling a large number of proposals during both the pre-processing and post-processing phases. To address this issue, DETR~\cite{detr} presents an end-to-end training framework that utilizes sparse queries. The DETR framework leverages the powerful transformer encoder-decoder architecture to encode global information and employs a set of queries to decode corresponding predictions, with the number of predictions being equal to the number of queries, typically set to 100 or 300. To further enhance the performance of DETR, the Deformable-DETR~\cite{deformdetr} introduces a multi-scale deformable attention module.

Sparse R-CNN~\cite{sparsercnn} and subsequent QueryInst~\cite{queryinst} are a hybrid of query-based detectors and two-stage detectors, offering the best of both worlds. It replaces the dense proposals of Faster R-CNN~\cite{fast-rcnn} or Mask R-CNN~\cite{mask-rcnn} with learnable query boxes and query features. The RoI features, obtained through query boxes, interact with corresponding query features in a cascade style, allowing for the final predictions to be decoded. The implementation of Sparse R-CNN not only accelerates the convergence speed but also improves the performance of query-based detectors.

Therefore, due to its better efficiency and performance, we choose QueryInst as our baseline model for our subsequent experiments. Meanwhile, we believe that other query-based detectors are capable as well.

%% file: 3-methods.tex
\begin{figure*}[htbp]
  \centering
  \includegraphics[width=0.95\linewidth]{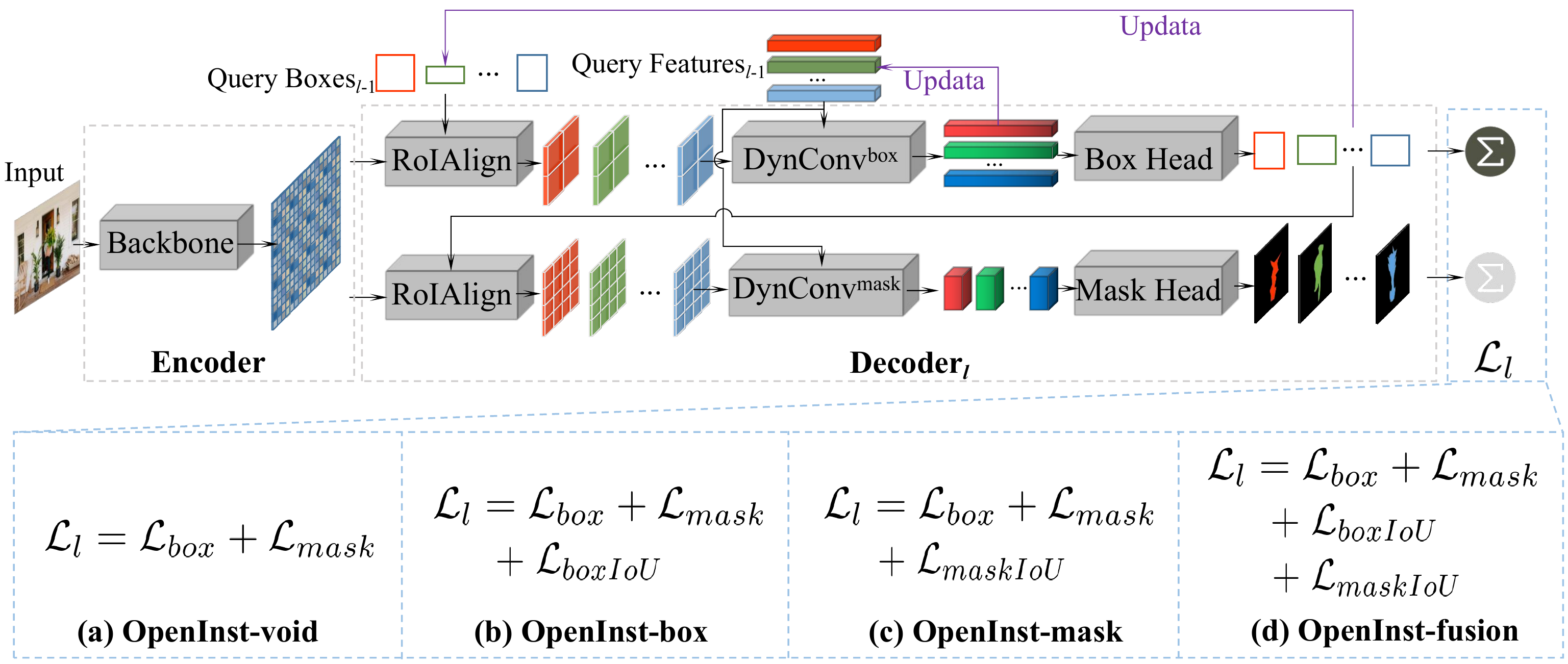}
  \centering
  \caption{Overview of OpenInst. OpenInst leverages the encoder-decoder architecture. The decoder contains L stages. Query boxes and features of the decoder are updated in each stage and serve as the queries for the next stage. According to the different learning objective and loss function, OpenInst can be categorized to 4 variants: OpenInst-void, OpenInst-box, OpenInst-mask, and OpenInst-fusion.}
  \label{fig:OpenInstvariants}
\end{figure*}


\subsection{Problem Definition} 
Open-world instance segmentation is to localize and segment all objects in arbitrary images. However, due to limited and fixed training images, it is unrealistic to expect the model to encounter all kinds of objects during training. Therefore, we use base categories $C_{base}$ and novel categories $C_{novel}$ to denote seen categories during training and unseen categories during testing, respectively. Testing images may contain both $C_{base}$ and $C_{novel}$ objects. There are two common settings for evaluating the performance of open-world instance segmentation models: cross-category and cross-dataset. In the cross-category setting, the model is trained with $C_{base}$ objects and tested on $C_{novel}$ objects. In cross-dataset setting, the model is trained on dataset $A$ and tested on dataset $B$, where dataset $B$ may contain both $C_{base}$ and $C_{novel}$ objects. The performance of both settings is measured by Average Recall (AR).

\subsection{Baseline Framework}
OpenInst can be constructed on many query-based frameworks. We adopt QueryInst~\cite{queryinst} as our baseline model for subsequent research. QueryInst is a typical query-based method for instance segmentation. Its decoder consists of six decode stages after the backbone and neck modules. Each stage is equipped with a unique dynamic box head as well as mask head. Query features are shared in both box and mask heads of each stage. Supervisions from ground truths are applied to all six stages. The training objective of each stage can be formulated as follows:
\begin{align}
\label{eq:level}\mathcal{L}_{l} &= \mathcal{L}_{objectness} + \mathcal{L}_{box} + \mathcal{L}_{mask}, \\[10pt]
\label{eq:obj:cls}\mathcal{L}_{objectness} &= \lambda_{cls} \cdot \mathrm{FL}(p_c, p_c^*), \\[10pt]
\label{eq:box}\mathcal{L}_{box} &= \lambda_{reg} \cdot \mathrm{L1}(t, t^*) + \lambda_{giou} \cdot \mathrm{GIoU}(t, t^*), \\[10pt]
\label{eq:mask}\mathcal{L}_{mask} &= \lambda_{mask} \cdot \mathrm{dice}(m, m^*).
\end{align}
$\mathcal{L}_{l}$ is the aggregation of all losses applied in the $l$ stage of the decoder. It consists of three losses: $\mathcal{L}_{objectness}$, $\mathcal{L}_{box}$ and $\mathcal{L}_{mask}$. $\mathrm{FL}$ denotes focal loss~\cite{focalloss}. $\mathcal{L}_{objectness}$ uses focal loss to learn the objectness as a classification score. $\mathcal{L}_{box}$ combines a $L1$ loss and a auxiliary $\mathrm{GIoU}$ loss~\cite{giou} for box regression learning. $\mathcal{L}_{mask}$ uses dice loss~\cite{diceloss} to learn the mask of an object. The $\lambda_{cls}$, $\lambda_{cls}$ and $\lambda_{cls}$ denote the weights of $\mathcal{L}_{objectness}$, $\mathcal{L}_{reg}$, $\mathcal{L}_{giou}$ and $\mathcal{L}_{mask}$ respectively. The $p_c$, $t$, and $m$ denote the classification score, box coordinates, and the mask of the predictions. The $p_c^*$, $t^*$, and $m^*$ denote the classification score, box coordinates, and the mask of the corresponding ground truth. The total loss $\mathcal{L}_{total}$ used optimize the detector is a sum of all $\mathcal{L}_{l}$:
\begin{equation}
    \mathcal{L}_{total} = \sum_{l=1}^{L} \mathcal{L}_{l},
\end{equation}
where $L$ denotes the number of layers in the decoder.

Query-based methods such as QueryInst predict all instances directly, with the number of predicted instances determined prior to training. The decoder refines \textbf{all} predictions stage by stage. 
When evaluating the generalizability of detectors using the recall metric, the objectness score can be omitted as the primary metric is often Average Recall (AR) at a budget of 100. This is the main metric used to evaluate the generalizability of open-world instance segmentation methods. To establish a fair comparison with prior works that do not use objectness scores, we set the query number to 100.

\subsection{OpenInst Framework}

Besides the difference in the complexity and performance, there is another notable difference in the utilization of the objectness score between dense proposal-based methods and query-based methods. 
In the case of dense proposal-based methods, the objectness score is employed in two ways. Firstly, it serves as a learning objective during the training phase and secondly, it functions as a ranking indicator in both the training and testing phases, enabling the filtering of boxes with high confidence from dense proposals. 
On the other hand, in query-based methods, the objectness score, which is mostly a classification score, is used only as an auxiliary learning objective in the training phase. It does \textbf{not} serve as a ranking index for filtering boxes from dense proposals, as the number of boxes in query-based methods is fixed. The existing boxes are refined stage by stage.

To investigate the effect of geometric cues on query-based methods, we have applied them to QueryInst and removed the classification branch, replacing it with either a box IoU branch or a mask IoU branch, resulting in the variants named OpenInst-box and OpenInst-mask, respectively. The variant that uses both the box and mask IoU branch is named OpenInst-fusion.
Furthermore, based on the observation that the objectness score is not needed in calculating the main metric AR@100, we question the necessity of treating the objectness score as a learning objective in query-based methods. we remove all objectness branches and name the resulting variant OpenInst-void.
The comprehensive framework of OpenInst and its variants is depicted in Fig.~\ref{fig:OpenInstvariants}. Notably, the modifications made to QueryInst are minimal, as our primary goal is to simplify the structure while maintaining high performance.

We use $L1$ loss as the supervision to train box or mask IoU branches. The $\mathcal{L}_{objectness}$ in Eq.~\ref{eq:level} is modified as Eq.~\ref{eq:OpenInst}. The symbols $p_b$ and $p_m$ denote the predicted box IoU score and mask IoU score, respectively. Meanwhile, the ground-truth scores are represented by $p_b^*$ and $p_m^*$. The symbols $\lambda_{box IoU}$ and $\lambda_{mask IoU}$ denote the corresponding weights used in the training phase.

\begin{equation}\label{eq:OpenInst}
\resizebox{\textwidth}{!}{%
$\mathcal{L}_{objectness} = \begin{cases}
0, &\text{if $\mathcal{L}_{objectness}$ in OpenInst-void} \\
\lambda_{box IoU} \cdot \mathrm{L1}(p_b, p_b^*), &\text{if $\mathcal{L}_{objectness}$ in OpenInst-box} \\
\lambda_{mask IoU} \cdot \mathrm{L1}(p_m, p_m^*), &\text{if $\mathcal{L}_{objectness}$ in OpenInst-mask} \\
\lambda_{box IoU} \cdot \mathrm{L1}(p_b, p_b^*) + \lambda_{mask IoU} \cdot \mathrm{L1}(p_m, p_m^*), &\text{if $\mathcal{L}_{objectness}$ in OpenInst-fusion}
\end{cases}$%
}
\end{equation}

\subsection{Best Practice in Open-World Instance Segmentation}
\textbf{Deformable modules.}
Deformable convnet~\cite{dcn,dcnv2} is a powerful tool in the closed-world object detection task as well as instance segmentation task. The deformable property makes it more precise than vanilla convnet. In the open-world instance segmentation task, we also want to leverage the deformable property to find more potential objects in an image. We add the advanced deformable convnet to our models and gain non-trivial promotions.

\textbf{Neck structure.}
Since FPN~\cite{fpn} has been raised, lots of works have concentrated on neck structure designing, such as BiFPN~\cite{bifpn}, PAFPN~\cite{pafpn}, and NAS-FPN~\cite{nasfpn}. Those works try to enhance the expressiveness of feature maps at different scales, in order to capture more accurate information from feature maps. These FPNs have made great contributions to the closed-world instance segmentation task. We here extend their applicability to the open-world instance segmentation, since the communication conducted in different feature maps may also enhance the information of novel objects. We choose BiFPN as the neck of our model.

%% file: 4-exp.tex
We have conducted adequate experiments in this section to reveal what kinds of objectness is beneficial for query-based methods in open-world instance segmentation task. 

\begin{table*}[htbp]
\centering
\caption{Results of cross-category generalizability evaluation on VOC$\to$Non-VOC scenario. $\mathrm{AR^{box}}$ denotes the box AR performance at a budget of 100. $\mathrm{AR}$ without superscript denotes mask AR performance at a budget of 100. \textbf{Bold} scores denote the best performance of each metric. OpenInst$^*$ denotes OpenInst trained with pseudo annotations produced by GGN.}
\label{tab:exp:voc}
\resizebox{\textwidth}{!} 
{
\begin{tabular}{lcccccccc}
\toprule
Methods & Ref & $\mathrm{AR^{box}}$ & $\mathrm{AR}$ & $\mathrm{AR_{0.5}}$ & $\mathrm{AR_{0.75}}$  & $\mathrm{AR_{small}}$ & $\mathrm{AR_{med}}$ & $\mathrm{AR_{large}}$ \\
\midrule
OLN~\cite{oln}        & ICRA22      & \textbf{33.0}          & 26.9          & -             & -                & -              & -             & -       \\
LDET~\cite{ldet}      & ECCV22      & 30.8          & 27.4          & -             & -                & -              & -             & -       \\
GGN~\cite{ggn}        & CVPR22      & 31.5          & 28.7          & -             & -                & -              & -             & -       \\
SOIS~\cite{sois}      & ARXIV22     &  -            & 11.0          &  -            &  -               &  4.9           &  9.2          & 24.8    \\
\textbf{OpenInst}     & -             & 32.0          & 28.2          & 44.8          & 29.4             & \textbf{11.6}           & 33.4          & 54.8    \\
\textbf{OpenInst$^*$} & -             & 33.0          & \textbf{30.1}          & \textbf{47.5}          & \textbf{31.9}             &  7.4           & \textbf{38.5}          & \textbf{64.2}    \\
\bottomrule
\end{tabular}
}
\end{table*}

\subsection{Datasets and Evaluation}
We conduct our experiments on six popular datasets: COCO~\cite{coco}, UVO~\cite{uvo}, Objects365~\cite{objects365}, Mapillary Vistas~\cite{mapillary}, LVIS~\cite{lvis}, and Cityscapes~\cite{cityscapes}, which are widely used in the closed-world instance segmentation task. LVIS, Mapillary Vistas, and Objects365 are only used for evaluation. Cityscapes is only used for training. COCO and UVO are used for both training and evaluation. 

\textbf{COCO} is a widely-used dataset in object detection and instance segmentation.
Following OLN and LDET, we use the train2017 split of COCO for training and the val2017 split for evaluation, which contain 117k and 5k images respectively. COCO covers 80 object categories, which are a superset of categories in the PASCAL VOC dataset.
We use COCO in both cross-category and cross-dataset generalizability evaluations. 
\textbf{UVO} is a class-agnostic and exhaustively labeled dataset. It is specially designed for the open-world instance segmentation task. We use UVO v0.5 for a fair comparison with other methods. UVO v0.5 contains 15315 images for training and 7356 images for validation. Following OLN, GGN, and LDET, we mainly use the validation split for cross-dataset generalizability evaluations. Following SOIS, we also conduct the so-called inner-dataset evaluation, which means using the training set of UVO v0.5 to train the model, and using the validation set of UVO v0.5 for evaluation. 
\textbf{Objects365} is large scale dataset for object detection and has 80k images for validation. Following LDET, we sample 5k images from the validation split of Objects365 for cross-dataset generalizability evaluation. Since Objects365 only has box-level annotations. We only apply open-world object detection evaluation on Objects365.
\textbf{Mapillary Vistas} is a street-centric dataset.
We use the validation split of Mapillary Vistas on version v1.2 for cross-dataset generalizability evaluation.
\textbf{LVIS} is a large-vocabulary dataset for instance segmentation. It contains more than 1200 categories. Following SOIS, we use the validation split (20k images) of LVIS for cross-dataset generalizability evaluation.
\textbf{Cityscapes} consists of urban scene images from 50 different cities. We use the 8 foreground classes from Cityscapes for training, and test the model on the validation split of Mapillary Vistas.

We use the average recall (AR) as the indicator to quantitatively measure the generalizability of different models. Since the AR metric is commonly used in many open-world instance segmentation works~\cite{oln,oln-fcos,ggn,ldet,sois}. Following OLN~\cite{oln}, GGN~\cite{ggn}, and LDET~\cite{ldet}, we also conduct experiments in two main settings: cross-category and cross-dataset.
AR@k denotes the average recall at a budget of k, which means only top k predictions of each image are used for calculating the average recall.

\textbf{Cross-category.}
We conduct cross-category generalizability evaluation only on one scenario: VOC$\to$Non-VOC. We only use annotations belonging to PASCAL VOC~\cite{pascalvoc} categories for supervision in the training phase. And we use annotations belonging to other categories for evaluation. In this experiment, we set the query number to 150. Because predictions belong to $C_{base}$ categories are excluded from the budget of the recall in evaluation.

\textbf{Cross-dataset.}
Cross-dataset generalizability evaluation uses two different datasets for training and testing respectively. The testing dataset contains both $C_{base}$ and $C_{novel}$ categories. We conduct cross-dataset generalizability evaluation on four scenarios: COCO$\to$UVO, COOC$\to$Objects365, COCO$\to$LVIS, and Cityscapes$\to$Mapillary Vistas.

\subsection{Implementations Details}
We build OpenInst on the powerful MMDetection~\cite{mmdet} library. We benchmark our method against the advanced QueryInst~\cite{queryinst} method. We use ResNet-50~\cite{resnet} as the backbone of our model, and leverage BiFPN~\cite{bifpn} instead of vanilla FPN~\cite{fpn} as the neck module. We use box IoU to displace the classification label as the learning objective unless otherwise specified. The box head is trained with L1 loss and GIoU loss~\cite{giou}, whose loss weights are set to 5.0 and 2.0 respectively. The mask head is trained with the dice loss. The number of decoder layer is set to 6 in all experiments.
We use the Adam~\cite{adam} optimizer as our solver with the learning rate set to 1e-4 and weight decay set to 5e-4. For \textbf{1x} configuration, we set the total epoch to 12, and make it decayed at epoch 8 and epoch 11 by 0.1 respectively. For \textbf{3x} configuration, the decay point is set to epoch 27 and epoch 33 respectively. We adopt RandomFlip as the only data augmentation method in our data pipeline for \textbf{1x} configuration and apply \textbf{L}arge \textbf{S}cale \textbf{J}ittor (LSJ) for \textbf{3x} configuration for fair comparison.

\begin{table*}[htbp]
\centering
\caption{Results of cross-category generalizability evaluation on COCO$\to$UVO scenario. $\mathrm{AR^{box}}$ denotes the box AR performance at a budget of 100. $\mathrm{AR}$ without superscript denotes the mask AR performance at a budget of 100. \textbf{Bold} scores denote the best performance of each metric. Aux. denotes that the method contains an auxiliary model and pseudo annotations.}
\label{tab:exp:uvo}
\resizebox{\textwidth}{!} 
{
\begin{tabular}{lcccccccccc}
\toprule
Methods & Ref & Aux. & Epochs & $\mathrm{AR^{box}}$ & $\mathrm{AR}$ & $\mathrm{AR_{0.5}}$ & $\mathrm{AR_{0.75}}$  & $\mathrm{AR_{small}}$ & $\mathrm{AR_{med}}$ & $\mathrm{AR_{large}}$ \\
\midrule
LDET~\cite{ldet}    & ECCV22  &             & 8  & 47.5          & 40.7          & -             & -                & 26.8           & 40.0           & 45.7   \\
GOOD~\cite{good}  & ICLR23    &  \checkmark & 8 & 50.3  & - & - & - & - & - & - \\
GGN~\cite{ggn}    & CVPR22    & \checkmark  & 8  & 52.8          & 43.4          & 71.7          & 44.5             & 23.3           & 44.4           & 50.0   \\
SOIS~\cite{sois}  & ARXIV22    & \checkmark  & 36 & -             & 51.3          & -             & -                & -              & -              & -      \\
\textbf{OpenInst} & -          &             & 12 & 59.1          & 48.7          & 72.6          & 51.4             & 26.4           & 44.3           & 60.4   \\
\textbf{OpenInst} & -          &             & 36 & \textbf{63.0} & \textbf{53.3} & \textbf{76.6} & \textbf{56.8}    & \textbf{31.8}  & \textbf{49.4}  & \textbf{64.3}   \\
\bottomrule
\end{tabular}
}
\end{table*}

\begin{table*}[htbp]
\centering
\caption{Results of cross-category generalizability evaluation on COCO$\to$Objects365 scenario. $\mathrm{AR^{box}}$ denotes the box AR performance at a budget of 100. \textbf{Bold} scores denote the best performance of each metric.}
\label{tab:exp:obj365}
\begin{tabular}{lcccccc}
\toprule
Methods & $\mathrm{AR^{box}}$ & $\mathrm{AR_{0.5}^{box}}$ & $\mathrm{AR_{0.75}^{box}}$  & $\mathrm{AR_{small}^{box}}$ & $\mathrm{AR_{med}^{box}}$ & $\mathrm{AR_{large}^{box}}$ \\
\midrule
Mask R-CNN~\cite{mask-rcnn}  & 38.5          & -             & -             & 24.0             & 40.1           & 50.2   \\
LDET~\cite{ldet}      & 41.1          & -             & -             & 26.1             & 43.8           & 52.8   \\
\textbf{OpenInst} & \textbf{50.1}          & 64.1          & 53.0          & \textbf{29.8}             & \textbf{51.8}           & \textbf{66.6}   \\
\bottomrule
\end{tabular}
\end{table*}

\begin{table}[htbp]
\centering
\caption{Results of cross-category generalizability evaluation on COCO$\to$LVIS scenario. $\mathrm{AR}$ denotes the mask AR performance at a budget of 100. \textbf{Bold} scores denote the best performance of each metric.}
\label{tab:exp:lvis}
\begin{tabular}{ccccccc}
\toprule
Methods & $\mathrm{AR}$ \\
\midrule
Mask R-CNN~\cite{mask-rcnn}    & 22.4  \\
LDET~\cite{ldet}        & 25.1  \\
Mask2Former~\cite{mask2former} & 24.5  \\
SOIS~\cite{sois}        & 25.2  \\
\textbf{OpenInst}   & \textbf{29.3}  \\
\bottomrule
\end{tabular}
\end{table}

\begin{table*}[htbp]
\centering
\caption{Results of cross-category generalizability evaluation on Cityscapes$\to$Mapillary Vistas scenario. $\mathrm{AR}$ denotes the mask AR performance at a budget of 100. \textbf{Bold} scores denote the best performance of each metric.}
\label{tab:exp:mapillary}
\begin{tabular}{ccccccc}
\toprule
Methods & $\mathrm{AR}$ & $\mathrm{AR_{0.5}}$ & $\mathrm{AR_{0.75}}$  & $\mathrm{AR_{small}}$ & $\mathrm{AR_{med}}$ & $\mathrm{AR_{large}}$ \\
\midrule
Mask R-CNN~\cite{mask-rcnn}      &  8.4          & 16.3          & -             & -                & -              & -      \\
LDET~\cite{ldet}      & 10.6          & \textbf{21.8}          & -             & -                & -              & -      \\
\textbf{OpenInst} & \textbf{11.6}          & 18.1          & 11.9          &  3.0             & 12.9           & 33.6   \\
\bottomrule
\end{tabular}
\end{table*}

\subsection{Cross-category Generalizability Evaluation}

\textbf{VOC}$\to$\textbf{Non-VOC.}
In the VOC$\to$Non-VOC scenario, we only concern with the performance of predictions matched with $C_{novel}$. Predictions matched with $C_{base}$ categories are not taken into account when calculating the recall scores. Therefore, we set the query number to 150 for a fair comparison with fellow works. The initialization is also changed from "Image Initialization" to "Random Initialization". Because "Image Initialization" focus more on $C_{base}$ objects in the refining process of the decoder, and thus weakens the localization performance on $C_{novel}$ objects. 
As shown in Fig.~\ref{tab:exp:voc}, OpenInst achieves comparable results against dense proposal-based methods~\cite{oln,ldet,ggn}. When compared with the query-based method SOIS, OpenInst obtains significant advantages. Powered by pseudo annotations, OpenInst$^*$ achieves state-of-the-art results on both boxes as well as mask AR and suppresses other methods by a notable margin. Besides, we have noticed that pseudo annotations mainly help improve the performance of medium and large-size novel objects, while downgrading the performance of small-size objects.

\subsection{Cross-dataset Evaluation}

\textbf{COCO}$\to$\textbf{UVO} is the most important scenario of cross-dataset generalizability evaluation. As shown in Tab.~\ref{tab:exp:uvo}, OpenInst achieves state-of-the-art results when trained with 12 epochs. OpenInst suppresses all dense proposal-based and query-based methods by a large margin. The mask AR score of OpenInst reaches 53.3, which is nearly 10 points higher than the advanced GGN~\cite{ggn}. When compared with query-based methods, OpenInst exceeds SOIS~\cite{sois} by 2 points on the mask AR score. Note that both GGN and SOIS have an auxiliary model and leverage pseudo annotations. Though an auxiliary model and pseudo annotations can enhance the detector, they also make the detector much heavier and the training pipeline more complicated. OpenInst possesses two advantages: better results and a simpler structure. LDET~\cite{ldet} is a dense proposal based without auxiliary models and pseudo annotation. Compared with LDET, we can find that the improvement of OpenInst trained with 12 epochs mainly comes from large objects. LDET suppress OpenInst trained with 12 epochs on $\mathrm{AR_{small}}$ by 0.4. We presume that query-based methods like OpenInst work better on large objects. Because the initialization of query boxes in this scenario is the size of the full images, which means query boxes and features are more likely to notice large objects. LDET leverages densely spread proposals of various sizes, which enables LDET to take objects of all sizes into account. When trained with 36 epochs with LSJ data augmentation, OpenInst obtains significant improvements on all size objects.

\textbf{COCO}$\to$\textbf{Objects365.}
Objects365 only has box-level annotations. We only use the box AR as the metric for evaluation. The original validation split of Objects365 contains 80k images. Following LDET~\cite{ldet}, we use the same subset of the original validation split for evaluation. The evaluation subset consists of 5k images. The taxonomy of Objects365 contains 365 categories and is a superset of the COCO taxonomy (80 categories). Objects365 contains all $C_{base}$ objects and lots of $C_{novel}$ objects. As shown in Tab.~\ref{tab:exp:obj365}, OpenInst outperforms LDET by a large margin. The box AR score of OpenInst reaches 50.1 and suppresses the score of LDET by 9.0 points. From the performance of the three object sizes, we can observe that the principal gap between the performance of LDET and OpenInst comes from large objects. This observation is consistent with that of the COCO$\to$UVO scenario.

\textbf{COCO}$\to$\textbf{LVIS} scenario is introduced by SOIS~\cite{sois}. LVIS is built upon COCO images but has more granular annotations. LVIS has 1203 categories while COCO has only 80. The taxonomy of LVIS is far bigger than that of COCO. Therefore, despite the overlapped images in the training and testing split, annotations of the overlapped images have a huge difference between the training split and the testing split. For those overlapped images, annotations from the training split do not help them cheat on the testing split. Results from COCO to LVIS can reveal the generalizability of the model all the same. As shown in Tab.~\ref{tab:exp:lvis}, OpenInst outperforms all available methods by at least 4.1 AR and achieves state-of-the-art results on the COCO$\to$LVIS scenario.

\textbf{Cityscapes}$\to$\textbf{Mapillary Vistas.}
Following LDET~\cite{ldet}, We train the detector using 8 object-level categories of Cityscapes: \textit{car, bicycles, motorcycle, train, bus, truck, person, rider}. For evaluation, we use the 35 foreground object-level categories of Mapillary Vistas. The number of categories in evaluation is four times as many as in training. As shown in Tab.~\ref{tab:exp:mapillary}, OpenInst achieves 29.3 mask AR@100, which promotes the performance of the query-based method SOIS by 4.1 AR.

\subsection{Ablation Study}

\begin{table}[htbp]
\centering
\caption{Results on class-agnostic dataset UVO. AR@100 and AR@10 denote mask AR performance at a budget of 100 and 10 respectively. \textbf{Bold} scores denote the best performance of each metric.}
\label{tab:exp:uvo2uvo}
\begin{tabular}{ccc}
\toprule
Methods & $\mathrm{AR@100}$ & $\mathrm{AR@10}$ \\
\midrule
Mask R-CNN~\cite{mask-rcnn}         & 22.8          & 20.0             \\
LDET~\cite{ldet}             & 35.6          & 23.7             \\
SOIS~\cite{sois}             & 41.9          & \textbf{29.2}             \\
\textbf{OpenInst}        & \textbf{43.1} & 20.8             \\
\bottomrule
\end{tabular}
\end{table}

\begin{table*}[htbp]
\centering
\caption{Effect of different geometric cues on COCO$\to$UVO scenario. $\mathrm{AR}^{box}$ denotes the box AR performance at a budget of 100. $\mathrm{AR}$ without superscript denotes the mask AR performance at a budget of 100. \textbf{Bold} scores denote the best performance of each metric. OpenInst-cls denotes the vanilla QueryInst trained in a class-agnostic way.}
\label{tab:exp:geocue}
\begin{tabular}{lccccccc}
\toprule
Methods & $\mathrm{AR^{box}}$ & $\mathrm{AR}$ & $\mathrm{AR_{0.5}}$ & $\mathrm{AR_{0.75}}$  & $\mathrm{AR_{small}}$ & $\mathrm{AR_{med}}$ & $\mathrm{AR_{large}}$ \\
\midrule
OpenInst-void   & 58.4          & 48.5          & 71.3          & 51.3          & 25.1             & 43.7           & \textbf{60.8}   \\
OpenInst-cls    & 55.7          & 44.9          & 72.1          & 46.7          & 24.0             & 41.8           & 55.2   \\
OpenInst-box    & \textbf{59.1} & \textbf{48.7} & \textbf{72.6} & \textbf{51.4} & \textbf{26.4}    & \textbf{44.3}  & 60.4   \\
OpenInst-mask   & 58.1          & 47.8          & 71.2          & 50.6          & 24.4             & 43.3           & 60.1 \\
OpenInst-fusion & 58.6          & 48.1          & 72.0          & 50.8          & 25.6             & 43.9           & 59.8   \\
\bottomrule
\end{tabular}
\end{table*}

\begin{table*}[htbp]
\centering
\caption{Effect of different modules on COCO$\to$UVO scenario. $\mathrm{AR}^{box}$ denotes the box AR performance at a budget of 100. $\mathrm{AR}$ without superscript denotes the mask AR performance at a budget of 100. \textbf{Bold} scores denote the best performance of each metric}
\label{tab:exp:module}
\resizebox{\textwidth}{!} 
{
\begin{tabular}{lcccccccc}
\toprule
DCN~\cite{dcnv2} & BiFPN~\cite{bifpn} & $\mathrm{AR^{box}}$ & $\mathrm{AR}$ & $\mathrm{AR_{0.5}}$ & $\mathrm{AR_{0.75}}$  & $\mathrm{AR_{small}}$ & $\mathrm{AR_{med}}$ & $\mathrm{AR_{large}}$ \\
\midrule
           &            & 56.3          & 46.4          & 71.3          & 48.9          & 24.3             & 42.2           & 57.8   \\
\checkmark &            & 57.9          & 47.9          & \textbf{72.8} & 50.5             & 25.3          & 43.8           & 59.4   \\
           & \checkmark & 58.4          & 47.8          & 71.4          & 50.5             & 25.8           & 43.4          & 59.3   \\
\checkmark & \checkmark & \textbf{59.1} & \textbf{48.7} & 72.6          & \textbf{51.4} & \textbf{26.4}    & \textbf{44.3}  & \textbf{60.4}   \\
\bottomrule
\end{tabular}
}
\end{table*}

\textbf{Open-world dataset evaluation.} Following SOIS~\cite{sois}, we conduct extra experiments on the open-world dataset UVO. We use the training split (15k images) and validation split (7856 images) of UVO for training and evaluation respectively. For a fair comparison, we set the image size to 640, to be consistent with SOIS. As shown in Tab.~\ref{tab:exp:uvo2uvo}, OpenInst achieves the best performance on the mask AR@100. OpenInst suppresses the prior query-based SOIS by 1.2 points. Whereas, OpenInst performs poorly on the mask AR@10 metric. The result of OpenInst on AR@10 indicates that using the box IoU as a ranking indicator is inadequate. Box IoU is not a good ranking indicator.

\textbf{Effect of geometric cues.} We use OpenInst-void as the baseline model. Based on this baseline model, we add the classification, box IoU, mask IoU, and the geometric mean of box and mask IoU as the learning objective respectively for training. The combination of OpenInst and classification is a vanilla QueryInst trained in a class-agnostic way. The remaining combinations are illustrated in Sec.~\ref{sec:method}. As shown in Tab.~\ref{tab:exp:geocue}, OpenInst-void has achieved impressive results. These results effectively demonstrate that \textbf{explicitly learning objectness is not crucial in the open-world localization and instance segmentation problem}, which is a highly insightful discovery. When classification is applied as a learning objective, the performances of box and mask AR drop by 2.7 and 3.6 respectively. The performance on $\mathrm{AR_{0.75}}$ experienced a noticeable decrease of 4.3 points. It can be inferred that the classification primarily resulted in insufficient fineness instead of accuracy in the predictions. The impacts of the box and mask IoU is minor than that of the classification. The box IoU has a positive effect on the generalizability while the mask IoU has a negative effect. We presume that learning the mask IoU is more difficult than learning box IoU. Because masks always have irregular shapes and boxes are always a rectangle. Being biased to the difficult mask IoU learning objective degrades the generalizability of the model. From Tab.~\ref{tab:exp:geocue} we can observe that only setting box IoU as the learning objective can slightly improve the performance. Query-based methods without learning any objectness have already been good detectors for the open-world instance segmentation task.

\textbf{Effect of DCN and BiFPN}. As shown in Tab.~\ref{tab:exp:module}, both DCN~\cite{dcnv2} and BiFPN~\cite{bifpn} have a positive effect on all metrics. The deformable mechanism of DCN expands the receptive field of the detector, making it larger and irregular. The detector can find more objects from a larger receptive field and output more precise predictions through irregular shapes. As can be seen in the second row of Tab.~\ref{tab:exp:module}, DCN improves the performance of $\mathrm{AR_{0.5}}$ by an increase of 1.5 points. In comparison, BiFPN brings even greater improvements to small objects. BiFPN designs a dedicated structure that makes full use of multi-scale feature maps. Feature maps with high resolution are enhanced, which makes it easier to locate small objects. After combining the two modules, the mask AR performance is further boosted to 48.7. This shows that DCN and BiFPN are complementary to each other in the open-world instance segmentation task.

%% file: 5-conclusion.tex
We propose OpenInst, a simple yet effective query-based method for open-world instance segmentation. Our method achieves state-of-the-art results on both cross-category and cross-dataset generalization evaluations, outperforming prior dense proposal-based and query-based methods while having a simpler structure. Our study shows that training query-based detectors solely with geometric information can lead to a robust open-world detector. Moreover, we discover that OpenInst performs well even without learning any objectness. We hope that OpenInst will serve as a strong baseline for future research and contribute to the open-world community.


%% file: OpenInst.bbl
\begin{thebibliography}{47}
\providecommand{\natexlab}[1]{#1}
\providecommand{\url}[1]{\texttt{#1}}
\expandafter\ifx\csname urlstyle\endcsname\relax
  \providecommand{\doi}[1]{doi: #1}\else
  \providecommand{\doi}{doi: \begingroup \urlstyle{rm}\Url}\fi

\bibitem[Alexe et~al.(2012)Alexe, Deselaers, and Ferrari]{measuringobj}
Bogdan Alexe, Thomas Deselaers, and Vittorio Ferrari.
\newblock Measuring the objectness of image windows.
\newblock \emph{IEEE transactions on pattern analysis and machine
  intelligence}, 34\penalty0 (11):\penalty0 2189--2202, 2012.

\bibitem[Carion et~al.(2020)Carion, Massa, Synnaeve, Usunier, Kirillov, and
  Zagoruyko]{detr}
Nicolas Carion, Francisco Massa, Gabriel Synnaeve, Nicolas Usunier, Alexander
  Kirillov, and Sergey Zagoruyko.
\newblock End-to-end object detection with transformers.
\newblock In Andrea Vedaldi, Horst Bischof, Thomas Brox, and Jan{-}Michael
  Frahm, editors, \emph{Computer Vision - {ECCV} 2020 - 16th European
  Conference, Glasgow, UK, August 23-28, 2020, Proceedings, Part {I}}, volume
  12346 of \emph{Lecture Notes in Computer Science}, pages 213--229. Springer,
  2020.

\bibitem[Chen et~al.(2019)Chen, Wang, Pang, Cao, Xiong, Li, Sun, Feng, Liu, Xu,
  et~al.]{mmdet}
Kai Chen, Jiaqi Wang, Jiangmiao Pang, Yuhang Cao, Yu~Xiong, Xiaoxiao Li,
  Shuyang Sun, Wansen Feng, Ziwei Liu, Jiarui Xu, et~al.
\newblock Mmdetection: Open mmlab detection toolbox and benchmark.
\newblock \emph{arXiv preprint arXiv:1906.07155}, 2019.

\bibitem[Cheng et~al.(2022)Cheng, Misra, Schwing, Kirillov, and
  Girdhar]{mask2former}
Bowen Cheng, Ishan Misra, Alexander~G. Schwing, Alexander Kirillov, and Rohit
  Girdhar.
\newblock Masked-attention mask transformer for universal image segmentation.
\newblock In \emph{{IEEE/CVF} Conference on Computer Vision and Pattern
  Recognition, {CVPR} 2022, New Orleans, LA, USA, June 18-24, 2022}, pages
  1280--1289. {IEEE}, 2022.

\bibitem[Cordts et~al.(2016)Cordts, Omran, Ramos, Rehfeld, Enzweiler, Benenson,
  Franke, Roth, and Schiele]{cityscapes}
Marius Cordts, Mohamed Omran, Sebastian Ramos, Timo Rehfeld, Markus Enzweiler,
  Rodrigo Benenson, Uwe Franke, Stefan Roth, and Bernt Schiele.
\newblock The cityscapes dataset for semantic urban scene understanding.
\newblock In \emph{2016 {IEEE} Conference on Computer Vision and Pattern
  Recognition, {CVPR} 2016, Las Vegas, NV, USA, June 27-30, 2016}, pages
  3213--3223. {IEEE} Computer Society, 2016.

\bibitem[Dai et~al.(2017)Dai, Qi, Xiong, Li, Zhang, Hu, and Wei]{dcn}
Jifeng Dai, Haozhi Qi, Yuwen Xiong, Yi~Li, Guodong Zhang, Han Hu, and Yichen
  Wei.
\newblock Deformable convolutional networks.
\newblock In \emph{{IEEE} International Conference on Computer Vision, {ICCV}
  2017, Venice, Italy, October 22-29, 2017}, pages 764--773. {IEEE} Computer
  Society, 2017.

\bibitem[Everingham et~al.(2015)Everingham, Eslami, Van~Gool, Williams, Winn,
  and Zisserman]{pascalvoc}
M.~Everingham, S.~M.~A. Eslami, L.~Van~Gool, C.~K.~I. Williams, J.~Winn, and
  A.~Zisserman.
\newblock The pascal visual object classes challenge: A retrospective.
\newblock \emph{International Journal of Computer Vision}, 111\penalty0
  (1):\penalty0 98--136, January 2015.

\bibitem[Fang et~al.(2021)Fang, Yang, Wang, Li, Fang, Shan, Feng, and
  Liu]{queryinst}
Yuxin Fang, Shusheng Yang, Xinggang Wang, Yu~Li, Chen Fang, Ying Shan, Bin
  Feng, and Wenyu Liu.
\newblock Instances as queries.
\newblock In \emph{2021 {IEEE/CVF} International Conference on Computer Vision,
  {ICCV} 2021, Montreal, QC, Canada, October 10-17, 2021}, pages 6890--6899.
  {IEEE}, 2021.

\bibitem[Ghiasi et~al.(2019)Ghiasi, Lin, and Le]{nasfpn}
Golnaz Ghiasi, Tsung{-}Yi Lin, and Quoc~V. Le.
\newblock {NAS-FPN:} learning scalable feature pyramid architecture for object
  detection.
\newblock In \emph{{IEEE} Conference on Computer Vision and Pattern
  Recognition, {CVPR} 2019, Long Beach, CA, USA, June 16-20, 2019}, pages
  7036--7045. Computer Vision Foundation / {IEEE}, 2019.

\bibitem[Girshick(2015)]{fast-rcnn}
Ross Girshick.
\newblock Fast r-cnn.
\newblock In \emph{Proceedings of the IEEE international conference on computer
  vision}, pages 1440--1448, 2015.

\bibitem[Girshick et~al.(2014)Girshick, Donahue, Darrell, and Malik]{rcnn}
Ross Girshick, Jeff Donahue, Trevor Darrell, and Jitendra Malik.
\newblock Rich feature hierarchies for accurate object detection and semantic
  segmentation.
\newblock In \emph{Proceedings of the IEEE conference on computer vision and
  pattern recognition}, pages 580--587, 2014.

\bibitem[Gu et~al.(2022)Gu, Lin, Kuo, and Cui]{ovvild}
Xiuye Gu, Tsung{-}Yi Lin, Weicheng Kuo, and Yin Cui.
\newblock Open-vocabulary object detection via vision and language knowledge
  distillation.
\newblock In \emph{The Tenth International Conference on Learning
  Representations, {ICLR} 2022, Virtual Event, April 25-29, 2022}.
  OpenReview.net, 2022.

\bibitem[Gupta et~al.(2019)Gupta, Dollar, and Girshick]{lvis}
Agrim Gupta, Piotr Dollar, and Ross Girshick.
\newblock Lvis: A dataset for large vocabulary instance segmentation.
\newblock In \emph{Proceedings of the IEEE/CVF conference on computer vision
  and pattern recognition}, pages 5356--5364, 2019.

\bibitem[Gupta et~al.(2021)Gupta, Narayan, Joseph, Khan, Khan, and
  Shah]{owdetr}
Akshita Gupta, Sanath Narayan, KJ~Joseph, Salman Khan, Fahad~Shahbaz Khan, and
  Mubarak Shah.
\newblock Ow-detr: Open-world detection transformer.
\newblock \emph{arXiv preprint arXiv:2112.01513}, 2021.

\bibitem[He et~al.(2016)He, Zhang, Ren, and Sun]{resnet}
Kaiming He, Xiangyu Zhang, Shaoqing Ren, and Jian Sun.
\newblock Deep residual learning for image recognition.
\newblock In \emph{2016 {IEEE} Conference on Computer Vision and Pattern
  Recognition, {CVPR} 2016, Las Vegas, NV, USA, June 27-30, 2016}, pages
  770--778. {IEEE} Computer Society, 2016.

\bibitem[He et~al.(2017)He, Gkioxari, Doll{\'a}r, and Girshick]{mask-rcnn}
Kaiming He, Georgia Gkioxari, Piotr Doll{\'a}r, and Ross Girshick.
\newblock Mask r-cnn.
\newblock In \emph{Proceedings of the IEEE international conference on computer
  vision}, pages 2961--2969, 2017.

\bibitem[Huang et~al.(2023)Huang, Geiger, and Zhang]{good}
Haiwen Huang, Andreas Geiger, and Dan Zhang.
\newblock {GOOD}: Exploring geometric cues for detecting objects in an open
  world.
\newblock In \emph{The Eleventh International Conference on Learning
  Representations}, 2023.
\newblock URL \url{https://openreview.net/forum?id=W-nZDQyuy8D}.

\bibitem[Huang et~al.(2019)Huang, Huang, Gong, Huang, and Wang]{msrcnn}
Zhaojin Huang, Lichao Huang, Yongchao Gong, Chang Huang, and Xinggang Wang.
\newblock Mask scoring r-cnn.
\newblock In \emph{IEEE Conference on Computer Vision and Pattern Recognition},
  2019.

\bibitem[Jia et~al.(2021)Jia, Yang, Xia, Chen, Parekh, Pham, Le, Sung, Li, and
  Duerig]{align}
Chao Jia, Yinfei Yang, Ye~Xia, Yi{-}Ting Chen, Zarana Parekh, Hieu Pham,
  Quoc~V. Le, Yun{-}Hsuan Sung, Zhen Li, and Tom Duerig.
\newblock Scaling up visual and vision-language representation learning with
  noisy text supervision.
\newblock In Marina Meila and Tong Zhang, editors, \emph{Proceedings of the
  38th International Conference on Machine Learning, {ICML} 2021, 18-24 July
  2021, Virtual Event}, volume 139 of \emph{Proceedings of Machine Learning
  Research}, pages 4904--4916. {PMLR}, 2021.

\bibitem[Jiang et~al.(2018)Jiang, Luo, Mao, Xiao, and Jiang]{iounet}
Borui Jiang, Ruixuan Luo, Jiayuan Mao, Tete Xiao, and Yuning Jiang.
\newblock Acquisition of localization confidence for accurate object detection.
\newblock In \emph{Proceedings of the European conference on computer vision
  (ECCV)}, pages 784--799, 2018.

\bibitem[Joseph et~al.(2021)Joseph, Khan, Khan, and Balasubramanian]{towod}
KJ~Joseph, Salman Khan, Fahad~Shahbaz Khan, and Vineeth~N Balasubramanian.
\newblock Towards open world object detection.
\newblock In \emph{Proceedings of the IEEE/CVF Conference on Computer Vision
  and Pattern Recognition}, pages 5830--5840, 2021.

\bibitem[Kim et~al.(2022)Kim, Lin, Angelova, Kweon, and Kuo]{oln}
Dahun Kim, Tsung-Yi Lin, Anelia Angelova, In~So Kweon, and Weicheng Kuo.
\newblock Learning open-world object proposals without learning to classify.
\newblock \emph{IEEE Robotics and Automation Letters (RA-L)}, 2022.

\bibitem[Kingma and Ba(2015)]{adam}
Diederik~P. Kingma and Jimmy Ba.
\newblock Adam: {A} method for stochastic optimization.
\newblock In Yoshua Bengio and Yann LeCun, editors, \emph{3rd International
  Conference on Learning Representations, {ICLR} 2015, San Diego, CA, USA, May
  7-9, 2015, Conference Track Proceedings}, 2015.

\bibitem[Konan et~al.(2022)Konan, Liang, and Yin]{oln-fcos}
Sachin Konan, Kevin~J Liang, and Li~Yin.
\newblock Extending one-stage detection with open-world proposals.
\newblock \emph{arXiv preprint arXiv:2201.02302}, 2022.

\bibitem[Lin et~al.(2014)Lin, Maire, Belongie, Hays, Perona, Ramanan,
  Doll{\'a}r, and Zitnick]{coco}
Tsung-Yi Lin, Michael Maire, Serge Belongie, James Hays, Pietro Perona, Deva
  Ramanan, Piotr Doll{\'a}r, and C~Lawrence Zitnick.
\newblock Microsoft coco: Common objects in context.
\newblock In \emph{European conference on computer vision}, pages 740--755.
  Springer, 2014.

\bibitem[Lin et~al.(2017{\natexlab{a}})Lin, Doll{\'{a}}r, Girshick, He,
  Hariharan, and Belongie]{fpn}
Tsung{-}Yi Lin, Piotr Doll{\'{a}}r, Ross~B. Girshick, Kaiming He, Bharath
  Hariharan, and Serge~J. Belongie.
\newblock Feature pyramid networks for object detection.
\newblock In \emph{2017 {IEEE} Conference on Computer Vision and Pattern
  Recognition, {CVPR} 2017, Honolulu, HI, USA, July 21-26, 2017}, pages
  936--944. {IEEE} Computer Society, 2017{\natexlab{a}}.

\bibitem[Lin et~al.(2017{\natexlab{b}})Lin, Goyal, Girshick, He, and
  Doll{\'{a}}r]{focalloss}
Tsung{-}Yi Lin, Priya Goyal, Ross~B. Girshick, Kaiming He, and Piotr
  Doll{\'{a}}r.
\newblock Focal loss for dense object detection.
\newblock In \emph{{IEEE} International Conference on Computer Vision, {ICCV}
  2017, Venice, Italy, October 22-29, 2017}, pages 2999--3007. {IEEE} Computer
  Society, 2017{\natexlab{b}}.

\bibitem[Liu et~al.(2018)Liu, Qi, Qin, Shi, and Jia]{pafpn}
Shu Liu, Lu~Qi, Haifang Qin, Jianping Shi, and Jiaya Jia.
\newblock Path aggregation network for instance segmentation.
\newblock In \emph{2018 {IEEE} Conference on Computer Vision and Pattern
  Recognition, {CVPR} 2018, Salt Lake City, UT, USA, June 18-22, 2018}, pages
  8759--8768. Computer Vision Foundation / {IEEE} Computer Society, 2018.

\bibitem[Mescheder et~al.(2019)Mescheder, Oechsle, Niemeyer, Nowozin, and
  Geiger]{occupancy}
Lars~M. Mescheder, Michael Oechsle, Michael Niemeyer, Sebastian Nowozin, and
  Andreas Geiger.
\newblock Occupancy networks: Learning 3d reconstruction in function space.
\newblock In \emph{{IEEE} Conference on Computer Vision and Pattern
  Recognition, {CVPR} 2019, Long Beach, CA, USA, June 16-20, 2019}, pages
  4460--4470. Computer Vision Foundation / {IEEE}, 2019.

\bibitem[Milletari et~al.(2016)Milletari, Navab, and Ahmadi]{diceloss}
Fausto Milletari, Nassir Navab, and Seyed{-}Ahmad Ahmadi.
\newblock V-net: Fully convolutional neural networks for volumetric medical
  image segmentation.
\newblock In \emph{Fourth International Conference on 3D Vision, 3DV 2016,
  Stanford, CA, USA, October 25-28, 2016}, pages 565--571. {IEEE} Computer
  Society, 2016.

\bibitem[Neuhold et~al.(2017)Neuhold, Ollmann, Bul{\`{o}}, and
  Kontschieder]{mapillary}
Gerhard Neuhold, Tobias Ollmann, Samuel~Rota Bul{\`{o}}, and Peter
  Kontschieder.
\newblock The mapillary vistas dataset for semantic understanding of street
  scenes.
\newblock In \emph{{IEEE} International Conference on Computer Vision, {ICCV}
  2017, Venice, Italy, October 22-29, 2017}, pages 5000--5009. {IEEE} Computer
  Society, 2017.

\bibitem[Radford et~al.(2021)Radford, Kim, Hallacy, Ramesh, Goh, Agarwal,
  Sastry, Askell, Mishkin, Clark, Krueger, and Sutskever]{clip}
Alec Radford, Jong~Wook Kim, Chris Hallacy, Aditya Ramesh, Gabriel Goh,
  Sandhini Agarwal, Girish Sastry, Amanda Askell, Pamela Mishkin, Jack Clark,
  Gretchen Krueger, and Ilya Sutskever.
\newblock Learning transferable visual models from natural language
  supervision.
\newblock In Marina Meila and Tong Zhang, editors, \emph{Proceedings of the
  38th International Conference on Machine Learning, {ICML} 2021, 18-24 July
  2021, Virtual Event}, volume 139 of \emph{Proceedings of Machine Learning
  Research}, pages 8748--8763. {PMLR}, 2021.

\bibitem[Ren et~al.(2015)Ren, He, Girshick, and Sun]{faster-rcnn}
Shaoqing Ren, Kaiming He, Ross Girshick, and Jian Sun.
\newblock Faster r-cnn: Towards real-time object detection with region proposal
  networks.
\newblock \emph{Advances in neural information processing systems}, 28, 2015.

\bibitem[Rezatofighi et~al.(2019)Rezatofighi, Tsoi, Gwak, Sadeghian, Reid, and
  Savarese]{giou}
Hamid Rezatofighi, Nathan Tsoi, JunYoung Gwak, Amir Sadeghian, Ian~D. Reid, and
  Silvio Savarese.
\newblock Generalized intersection over union: {A} metric and a loss for
  bounding box regression.
\newblock In \emph{{IEEE} Conference on Computer Vision and Pattern
  Recognition, {CVPR} 2019, Long Beach, CA, USA, June 16-20, 2019}, pages
  658--666. Computer Vision Foundation / {IEEE}, 2019.

\bibitem[Saito et~al.(2022)Saito, Hu, Darrell, and Saenko]{ldet}
Kuniaki Saito, Ping Hu, Trevor Darrell, and Kate Saenko.
\newblock Learning to detect every thing in an open world.
\newblock In \emph{European Conference on Computer Vision}, pages 268--284.
  Springer, 2022.

\bibitem[Shao et~al.(2019)Shao, Li, Zhang, Peng, Yu, Zhang, Li, and
  Sun]{objects365}
Shuai Shao, Zeming Li, Tianyuan Zhang, Chao Peng, Gang Yu, Xiangyu Zhang, Jing
  Li, and Jian Sun.
\newblock Objects365: A large-scale, high-quality dataset for object detection.
\newblock In \emph{Proceedings of the IEEE/CVF international conference on
  computer vision}, pages 8430--8439, 2019.

\bibitem[Sun et~al.(2021)Sun, Zhang, Jiang, Kong, Xu, Zhan, Tomizuka, Li, Yuan,
  Wang, and Luo]{sparsercnn}
Peize Sun, Rufeng Zhang, Yi~Jiang, Tao Kong, Chenfeng Xu, Wei Zhan, Masayoshi
  Tomizuka, Lei Li, Zehuan Yuan, Changhu Wang, and Ping Luo.
\newblock Sparse {R-CNN:} end-to-end object detection with learnable proposals.
\newblock In \emph{{IEEE} Conference on Computer Vision and Pattern
  Recognition, {CVPR} 2021, virtual, June 19-25, 2021}, pages 14454--14463.
  Computer Vision Foundation / {IEEE}, 2021.

\bibitem[Tan et~al.(2020)Tan, Pang, and Le]{bifpn}
Mingxing Tan, Ruoming Pang, and Quoc~V. Le.
\newblock Efficientdet: Scalable and efficient object detection.
\newblock In \emph{2020 {IEEE/CVF} Conference on Computer Vision and Pattern
  Recognition, {CVPR} 2020, Seattle, WA, USA, June 13-19, 2020}, pages
  10778--10787. Computer Vision Foundation / {IEEE}, 2020.

\bibitem[Tian et~al.(2019)Tian, Shen, Chen, and He]{fcos}
Zhi Tian, Chunhua Shen, Hao Chen, and Tong He.
\newblock Fcos: Fully convolutional one-stage object detection.
\newblock In \emph{Proceedings of the IEEE/CVF international conference on
  computer vision}, pages 9627--9636, 2019.

\bibitem[Uijlings et~al.(2013)Uijlings, Van De~Sande, Gevers, and
  Smeulders]{ss}
Jasper~RR Uijlings, Koen~EA Van De~Sande, Theo Gevers, and Arnold~WM Smeulders.
\newblock Selective search for object recognition.
\newblock \emph{International journal of computer vision}, 104\penalty0
  (2):\penalty0 154--171, 2013.

\bibitem[Wang et~al.(2021)Wang, Feiszli, Wang, and Tran]{uvo}
Weiyao Wang, Matt Feiszli, Heng Wang, and Du~Tran.
\newblock Unidentified video objects: A benchmark for dense, open-world
  segmentation.
\newblock In \emph{Proceedings of the IEEE/CVF International Conference on
  Computer Vision}, pages 10776--10785, 2021.

\bibitem[Wang et~al.(2022)Wang, Feiszli, Wang, Malik, and Tran]{ggn}
Weiyao Wang, Matt Feiszli, Heng Wang, Jitendra Malik, and Du~Tran.
\newblock Open-world instance segmentation: Exploiting pseudo ground truth from
  learned pairwise affinity.
\newblock In \emph{Proceedings of the IEEE/CVF Conference on Computer Vision
  and Pattern Recognition}, pages 4422--4432, 2022.

\bibitem[Xue et~al.(2022)Xue, Yu, Liu, Liu, Li, Yuan, Song, and Shou]{sois}
Xizhe Xue, Dongdong Yu, Lingqiao Liu, Yu~Liu, Ying Li, Zehuan Yuan, Ping Song,
  and Mike~Zheng Shou.
\newblock Single-stage open-world instance segmentation with cross-task
  consistency regularization.
\newblock \emph{CoRR}, abs/2208.09023, 2022.

\bibitem[Zang et~al.(2022)Zang, Li, Zhou, Huang, and Loy]{ovcm}
Yuhang Zang, Wei Li, Kaiyang Zhou, Chen Huang, and Chen~Change Loy.
\newblock Open-vocabulary detr with conditional matching.
\newblock In \emph{Computer Vision--ECCV 2022: 17th European Conference, Tel
  Aviv, Israel, October 23--27, 2022, Proceedings, Part IX}, pages 106--122.
  Springer, 2022.

\bibitem[Zhu et~al.(2019)Zhu, Hu, Lin, and Dai]{dcnv2}
Xizhou Zhu, Han Hu, Stephen Lin, and Jifeng Dai.
\newblock Deformable convnets {V2:} more deformable, better results.
\newblock In \emph{{IEEE} Conference on Computer Vision and Pattern
  Recognition, {CVPR} 2019, Long Beach, CA, USA, June 16-20, 2019}, pages
  9308--9316. Computer Vision Foundation / {IEEE}, 2019.

\bibitem[Zhu et~al.(2021)Zhu, Su, Lu, Li, Wang, and Dai]{deformdetr}
Xizhou Zhu, Weijie Su, Lewei Lu, Bin Li, Xiaogang Wang, and Jifeng Dai.
\newblock Deformable {DETR:} deformable transformers for end-to-end object
  detection.
\newblock In \emph{9th International Conference on Learning Representations,
  {ICLR} 2021, Virtual Event, Austria, May 3-7, 2021}. OpenReview.net, 2021.

\bibitem[Zitnick and Doll{\'a}r(2014)]{edgebox}
C~Lawrence Zitnick and Piotr Doll{\'a}r.
\newblock Edge boxes: Locating object proposals from edges.
\newblock In \emph{European conference on computer vision}, pages 391--405.
  Springer, 2014.

\end{thebibliography}
